\documentclass{article}

\usepackage{microtype}
\usepackage{graphicx}
\usepackage{subfigure}
\usepackage{booktabs} % for professional tables

\usepackage{hyperref}
\usepackage[accepted]{icml2023}

\usepackage{amsmath}
\usepackage{amssymb}
\usepackage{mathtools}
\usepackage{amsthm}
\usepackage{url}
\usepackage{color}
\usepackage{url}
\usepackage{verbatim}
\usepackage{graphicx}
\usepackage{booktabs}
\usepackage{multirow}
\usepackage{algorithmic}
\usepackage{algorithm}
\usepackage{array}
\usepackage{subfigure}
\usepackage{natbib}
\usepackage{textcomp}
\usepackage{stfloats}
\usepackage{bm}
\usepackage[capitalize,noabbrev]{cleveref}

\theoremstyle{plain}

\theoremstyle{definition}

\theoremstyle{remark}

\usepackage[textsize=tiny]{todonotes}

\newcommand{\haichao}[1]{\textcolor{black}{#1}}
\newcommand{\cui}[1]{\textcolor{black}{#1}}

\usepackage{colortbl}
\usepackage{tabularray}
\definecolor{mygray}{gray}{.9}
\usepackage{xcolor}
\usepackage{adjustbox}

\usepackage[capitalize]{cleveref}
\crefname{section}{Sec.}{Secs.}
\Crefname{section}{Section}{Sections}
\Crefname{table}{Table}{Tables}
\crefname{table}{Tab.}{Tabs.}

\icmltitlerunning{Learning Dynamic Query Combinations}

\begin{document}

\twocolumn[
\icmltitle{Learning Dynamic Query Combinations for Transformer-based Object Detection and Segmentation}

% It is OKAY to include author information, even for blind
% submissions: the style file will automatically remove it for you
% unless you've provided the [accepted] option to the icml2023
% package.

% List of affiliations: The first argument should be a (short)
% identifier you will use later to specify author affiliations
% Academic affiliations should list Department, University, City, Region, Country
% Industry affiliations should list Company, City, Region, Country

% You can specify symbols, otherwise they are numbered in order.
% Ideally, you should not use this facility. Affiliations will be numbered
% in order of appearance and this is the preferred way.
% \icmlsetsymbol{equal}{*}

% \begin{icmlauthorlist}
% \icmlauthor{Firstname1 Lastname1}{equal,yyy}
% \icmlauthor{Firstname2 Lastname2}{equal,yyy,comp}
% \icmlauthor{Firstname3 Lastname3}{comp}
% \icmlauthor{Firstname4 Lastname4}{sch}
% \icmlauthor{Firstname5 Lastname5}{yyy}
% \icmlauthor{Firstname6 Lastname6}{sch,yyy,comp}
% \icmlauthor{Firstname7 Lastname7}{comp}
% %\icmlauthor{}{sch}
% \icmlauthor{Firstname8 Lastname8}{sch}
% \icmlauthor{Firstname8 Lastname8}{yyy,comp}
% %\icmlauthor{}{sch}
% %\icmlauthor{}{sch}
% \end{icmlauthorlist}

% \icmlaffiliation{yyy}{Department of XXX, University of YYY, Location, Country}
% \icmlaffiliation{comp}{Company Name, Location, Country}
% \icmlaffiliation{sch}{School of ZZZ, Institute of WWW, Location, Country}

\begin{icmlauthorlist}
\icmlauthor{Yiming Cui}{uf,bd}
\icmlauthor{Linjie Yang}{bd}
\icmlauthor{Haichao Yu}{bd}
%\icmlauthor{}{sch}
%\icmlauthor{}{sch}
\end{icmlauthorlist}

\icmlaffiliation{uf}{Department of Electrical and Computer Engineering, University of Florida, Gainesville, USA}
\icmlaffiliation{bd}{ByteDance Inc., San Jose, USA}

\icmlcorrespondingauthor{Linjie Yang}{yljatthu@gmail.com}
% \icmlcorrespondingauthor{Firstname2 Lastname2}{first2.last2@www.uk}

% You may provide any keywords that you
% find helpful for describing your paper; these are used to populate
% the "keywords" metadata in the PDF but will not be shown in the document
% \icmlkeywords{Machine Learning, ICML}

\vskip 0.3in
]

% this must go after the closing bracket ] following \twocolumn[ ...

% This command actually creates the footnote in the first column
% listing the affiliations and the copyright notice.
% The command takes one argument, which is text to display at the start of the footnote.
% The \icmlEqualContribution command is standard text for equal contribution.
% Remove it (just {}) if you do not need this facility.

\printAffiliationsAndNotice{}  % leave blank if no need to mention equal contribution
% \printAffiliationsAndNotice{\icmlEqualContribution} % otherwise use the standard text.

\begin{abstract}
Transformer-based detection and segmentation methods use a list of learned detection queries to retrieve information from the transformer network and learn to predict the location and category of one specific object from each query. We empirically find that random convex combinations of the learned queries are still good for the corresponding models. We then propose to learn a convex combination with dynamic coefficients based on the high-level semantics of the image. The generated dynamic queries, named modulated queries, better capture the prior of object locations and categories in the different images. Equipped with our modulated queries, a wide range of DETR-based models achieve consistent and superior performance across multiple tasks including object detection, instance segmentation, panoptic segmentation, and video instance segmentation. 
\end{abstract}

\section{Introduction} 
Object detection is a fundamental yet challenging task in computer vision, which aims to localize and categorize objects of interest in the images simultaneously. Traditional detection models ~\citep{ren2015faster, Cai_2019, duan2019centernet, lin2017focal, lin2017feature} use complicated anchor designs and heavy post-processing steps such as Non-Maximum-Suppression (NMS) to remove duplicated detections. Recently, Transformer-based object detectors such as DETR \citep{detr} have been introduced to simplify the process. In detail, DETR combines convolutional neural networks (CNNs) with Transformer \citep{vaswani2017attention} by introducing an encoder-decoder framework to generate a series of predictions from a list of object queries. Following works improve the efficiency and convergence speed of DETR with modifications to the attention module~\citep{zhu2021deformable, roh2021sparse}, and divide queries into positional and content queries~\citep{liu2022dabdetr,meng2021-CondDETR, wang2022anchor, zhang2022dino}. This paradigm is also adopted for instance/panoptic segmentation, where each query is associated with one specific object mask in the decoding stage of the segmentation model~\citep{cheng2021mask2former, dong2021solq, cheng2021per, hu2021istr, wang2021end}.

\begin{figure} [!bt]
    \centering
    \includegraphics[width=8.5cm]{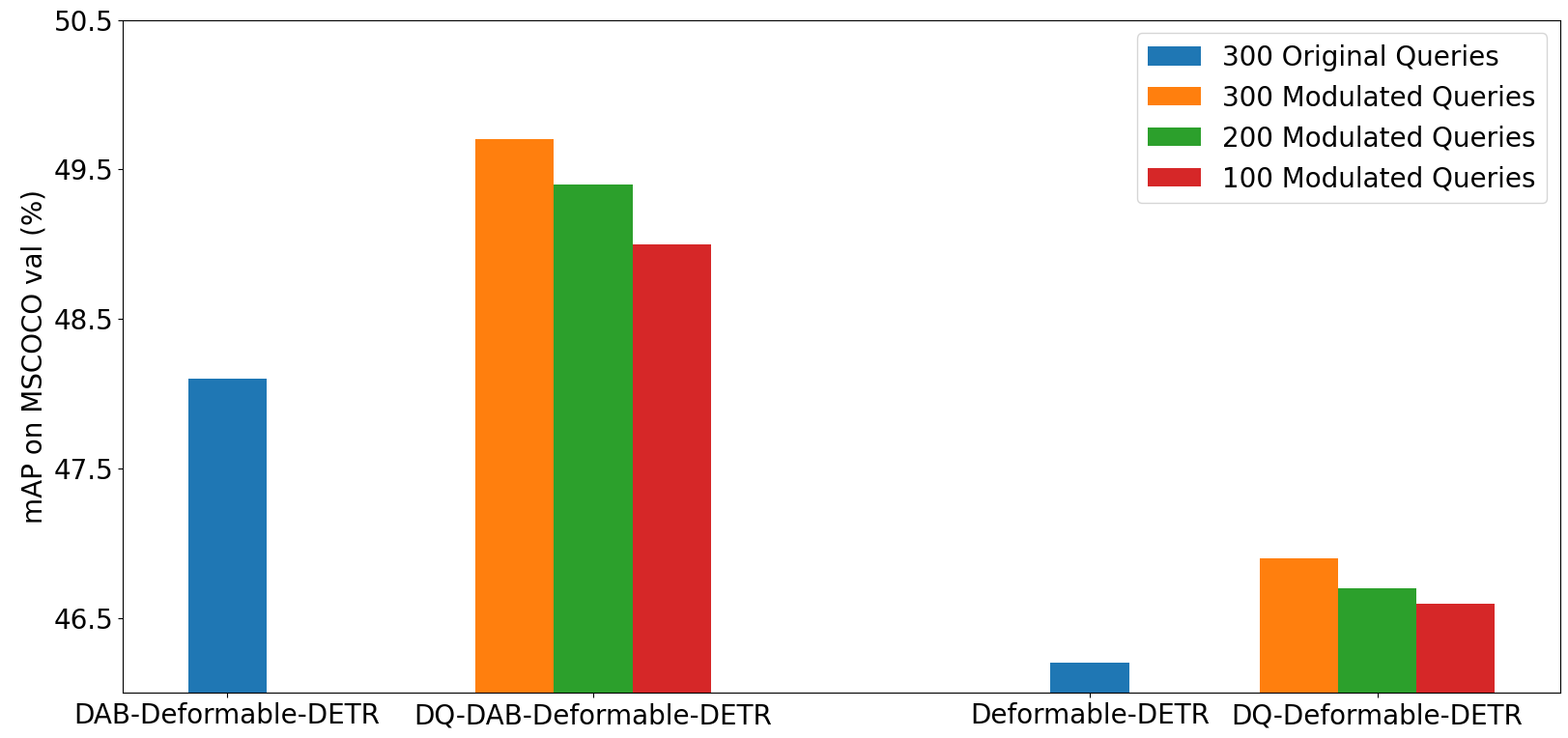}
    \vspace{-0.6cm}
    \caption{Comparison of DETR-based detection models integrated with and without our methods on MS COCO \citep{lin2014microsoft} \texttt{val} benchmark. ResNet-50 \citep{he2016deep} is used as the backbone. }
     \vspace{-0.4cm}
    \label{fig: resultCompare}
\end{figure}
The existing DETR-based detection models always use a list of fixed queries, regardless of the input image. The queries will attend to different objects in the image through a multi-stage attention process. Here, the queries serve as global priors for the location and semantics of target objects in the image. In this paper, we would like to associate the detection queries with the content of the image, i.e., adjusting detection queries based on the high-level semantics of the image in order to capture the distribution of object locations and categories in this specific scene. For example, when the high-level semantics shows the image is a group photo, we know that there will be a group of people (category) inside the image and they are more likely to be close to the center of the image (location).
 
Since the detection queries are implicit features that do not directly relate to specific locations and object categories in the DETR framework, it is hard to design a mechanism to change the queries while keeping \cui{them} within a meaningful ``query'' subspace to the model. Through an empirical study, we notice that convex combinations of learned queries are still good queries to different DETR-based models, achieving similar performance as the originally learned queries (See Section 3.2). Motivated by this, we propose a method to generate dynamic detection queries, named \emph{modulated queries}, based on the high-level semantics of the image in DETR-based methods while constraining the generated queries in a sequence of convex hulls spanned by the static queries. Therefore, the generated detection queries are more related to the target objects in the image and stay in a meaningful subspace. We show the superior performance of our approach combined with a wide range of DETR-based models on MS COCO \citep{lin2014microsoft}, CityScapes \citep{Cordts2016Cityscapes} and YouTube-VIS \citep{yang2019video} benchmarks with multiple tasks, including object detection, instance segmentation, and panoptic segmentation. 
In Figure~\ref{fig: resultCompare}, we show the performance of our method on object detection combined with two baseline models. When integrated with our proposed method, the mAP of recent detection models DAB-Deformable-DETR~\citep{liu2022dabdetr} can be increased by $1.6\%$. \textcolor{black}{With fewer modulated queries, our method can still achieve better performance than baseline models on both Deformable-DETR and DAB-Deformable-DETR.}

\section{Related Works}
\noindent\textbf{Transformers for object detection.} Traditional CNN-based object detectors require manually designed components such as anchors \citep{ren2015faster, Cai_2019, girshick2015fast, he2017mask} or post-processing steps such as NMS\citep{neubeck2006efficient, hosang2017learning, rothe2015non}. Transformer-based detectors directly generate predictions for a list of target objects with a series of learnable queries. Among them, DETR \citep{detr} first combines the sequence-to-sequence framework with learnable queries and CNN features for object detection.

Following DETR, multiple works \citep{chen2022group, zhu2021deformable, roh2021sparse, jia2022detrs, zhang2022dino, liu2022dabdetr} were proposed to improve its convergence speed and accuracy. Deformable-DETR \citep{zhu2021deformable} and Sparse-DETR \citep{roh2021sparse} replace the self-attention modules with more efficient attention operations where only a small set of key-value pairs are used for calculation. Conditional-DETR \citep{tian2020conditional} changes the queries in DETR to be conditional spatial queries, which speeds up the convergence process. SMCA-DETR \citep{gao2021fast} introduces pre-defined Gaussian maps around the reference points. Anchor-DETR \citep{wang2022anchor} generates the object queries using anchor points rather than a set of learnable embeddings. DAB-DETR \citep{liu2022dabdetr} directly uses learnable box coordinates as queries which can be refined in the Transformer decoder layers. DN-DETR \citep{li2022dn} improves the convergence speed of DETR by introducing noises to the ground truths and forcing the Transformer decoder to reconstruct the bounding boxes. DINO \citep{zhang2022dino} and DN-DETR \cite{li2022dn} introduce a strategy to train models with noisy ground truths to help the model learn the representation of the positive samples more
efficiently. 

Recently, Group-DETR \cite{chen2022group} and HDETR \cite{jia2022detrs} both added auxiliary queries and a one-to-many matching loss to improve the convergence of the DETR-based models. They still use static queries which does not change the general architecture of DETR. All these Transformer-based detection methods use fixed initial detection queries learned on the whole dataset. he queries will attend to different objects in the image through a multi-stage attention process. Without the global context, the queries might attend to regions that do not contain any objects or search for categories that do not exist in the image, which may limit the model's performance. In contrast, we propose to modulate the queries based on the image's content, which generates more effective queries for the current image. 

\noindent\textbf{Transformers for object segmentation.} Besides object detection, Transformer-based models are also proposed for object segmentation tasks including image instance segmentation \citep{he2017mask, wang2020solo, yolact-iccv2019, wang2020solov2, yolact-plus-tpami2020, cao2020sipmask}, panoptic segmentation  \textcolor{black}{\citep{kirillov2019panoptic, wang2021max, zhang2021k, xiong2019upsnet}} and video instance segmentation (VIS) \citep{yang2019video, hwang2021video, liu2021sg, liu2019spatio}. In  DETR \citep{detr}, a mask head is introduced on top of the decoder outputs to generate the predictions for panoptic segmentation. Following DETR, ISTR \citep{hu2021istr} generates low-dimensional mask embeddings, which are matched with the ground truth mask embeddings using Hungarian Algorithm for instance segmentation. 
SOLQ \citep{dong2021solq} uses a unified query representation for class, location, and mask.

Besides image object segmentation, researchers have begun to investigate object segmentation in video domains \citep{wang2021end, wu2022seqformer, thawakar2022video, yang2022temporally, hwang2021video}. VisTR \citep{wang2021end} extends DETR from the image domain to the video domain by introducing an instance sequence matching and segmentation pipeline for video instance segmentation. SeqFormer \citep{wu2022seqformer} utilizes video-level instance queries where each query attends to a specific object across frames in the video. MSSTS-VIS \citep{thawakar2022video} introduces a multi-scale spatial-temporal split attention module for video instance segmentation.

Recently, multiple works \citep{cheng2021mask2former, jain2022oneformer, cheng2021per, liang2023clustseg} pay attention to unified frameworks for object segmentation tasks in both image and video domains. \citet{cheng2021per} present MaskFormer, a straightforward mask classification model. It predicts binary masks linked to global class labels, simplifying semantic and panoptic segmentation tasks and achieving impressive empirical outcomes. By extending MaskFormer, Mask2Former \citep{cheng2021mask2former} introduces masked attention to extract localized features and predict output for panoptic, instance, and semantic segmentation in a unified framework. These Transformer-based models follow the general paradigm of DETR and use fixed queries regardless of the input.

\begin{table*}[!bt]
    \centering 
    \begin{tabular}{l|l|l|l|l|l}
    \toprule
    \multirow{2}{*}{Model} & \multicolumn{2}{l|}{DAB-DETR} & \multicolumn{2}{l|}{Deformable-DETR} &  Mask2Former\\
    \cline{2-6}
    \rule{0pt}{10pt}
    & $r=2$ & $r=4$ & $r=2$ & $r=4$ & $r=2$\\
    \midrule
    Convex Combination & 37.9$(\pm{0.10})$ & 30.4$(\pm{0.20)}$ & 35.0$(\pm{0.20})$ & 24.2$(\pm{0.05})$&  41.2$(\pm 0.10)$ \\
    Non-convex Combination & 37.0$(\pm{0.10})$ & 29.5$(\pm{0.10})$ &  32.6$(\pm0.25)$ & 24.0$(\pm{0.10})$ & 40.7$(\pm{0.45})$\\
    Averaged Combination & 37.0 & 28.4 & 32.9 & 22.5  & 40.9\\
    Queries sampled randomly & 39.7$(\pm{0.05})$ & 33.9$(\pm{0.15})$ &  39.8$({\pm0.30})$ & 28.1$(\pm{0.30})$ & 41.7$(\pm{0.10})$ \\
    \bottomrule
    \end{tabular}
        \vspace{-0.3cm}
    \caption{Comparison of pretrained detection models DAB-DETR \citep{liu2022dabdetr} and Deformable-DETR and segmentation model Mask2Former \citep{cheng2021mask2former} with different queries. The shown metrics are box mAP for detection and mask mAP for segmentation. ResNet-50 is used as the backbone and models are evaluated on MS COCO \texttt{val}.}
     \vspace{-0.4cm}
    \label{tab: convex}
\end{table*}

\noindent\textbf{Dynamic deep neural networks.} Dynamic deep neural network~\citep{han2021dynamic} aims at adjusting the computation procedure of a neural network adaptively in order to reduce the overall computation cost or enhance the model capacity. Slimmable networks \citep{yu2018slimmable, yu2019autoslim, li2021dynamic} introduce a strategy to adapt to multiple devices by simply changing channel numbers without the need for retraining. Dynamic Convolution \citep{chen2020dynamic} proposes a dynamic perceptron that uses dynamic attention weights to aggregate multiple convolution kernels based on the input features. Similar to dynamic convolution, CondConv \citep{yang2019condconv} introduces an operation named conditionally parameterized convolutions, which learns specialized convolutional kernels for each individual input. 

On object detection, Dynamic R-CNN \citep{zhang2020dynamic} proposes a new training strategy to dynamically adjust the label assignment for two-stage object detectors based on the statics of proposals. Dynamic-DETR \citep{li2021dynamic} introduces a dynamic attention module to DETR that dynamically adjusts attention according to factors such as the importance of scale to improve the performance on small objects and convergence speed. \citet{cui2022dynamic} proposes to train a single detection model which can adjust the number of proposals based on the complexity of the input image. DFA \citep{cui2022dynamic2, Cui_2023_CVPR} simplifies the feature aggregation process for video object detection by using a dynamic number of frames to enhance the object representations. \citet{wang2021not} introduces a Dynamic Transformer to determine the number of tokens according to the input image for efficient image recognition, by stacking multiple Transformer layers with increasing numbers of tokens. SODAR \citep{wang2021sodar} focuses on instance segmentation based on a one-stage SOLO model \citep{wang2020solo, wang2020solov2} for better performance. It improves the final segmentation quality by leveraging the rich neighboring information with a learning-based aggregation method. This model cannot be directly applied to other models, such as DETR-based models. GCNet \citep{cao2019gcnet} is designed for long-range dependency modeling in traditional convolutional networks. It simplifies the Non-Local Network (NLNet) by only considering the global context in the attention block. 

Both SODAR and GCNet deal with CNN-based model backbones, which are different from the Transformer encoder-decoder structure in the DETR framework. We believe our method can shed light on dynamic model designing in the Transformer paradigm. In contrast to the existing work, we explore \cui{generating} dynamic queries  for a wide range of DETR-based models using the same framework. Our focus is not to reduce the computation cost of DETR-based models, but to improve the model performances with queries more related to the content of each individual image.

\section{Methodology}
\subsection{Preliminary}
% DETR \citep{detr} is the milestone of most of the existing query-based object detectors and we revisit its framework as follows:

% Given the feature maps $x\in\mathbb{R}^{H\times W\times C}$ after the backbone, DETR uses a sequence-to-sequence architecture to project $x$ into the features of a set of object queries. Then, a multiple-layer feed-forward neural network is used to transform the features of object queries to object categories and corresponding bounding boxes, which are finally matched with the ground truths. In the encoder part of the sequence-to-sequence framework, every pixel in the feature maps $x$ is used as both query and key elements for multi-head self-attention calculation. In the decoder part, 

% $N_q$ randomly initialized queries $Q\in\mathbb{R}^{N_q\times f}$ are fed into the decoder network $\mathcal{N}_{dec}$ together with $F$ to generate the corresponding bounding box $B$ and object categories $C$
% \begin{equation}
%     B, C = \mathcal{N}_{dec}\left(Q, F\right)
% \end{equation}
% where $f$ represents the number of channels. 

We first summarize the inference process of the existing Transformer-based models for a series of tasks, including object detection, instance segmentation, and panoptic segmentation, as the following Equation:
\begin{equation}
\begin{aligned}
    \bm{Y} &= \mathcal{N}_{t} \left(\mathcal{N}_{dec}\left(\mathcal{N}_{enc}\left(\bm{F}\right), \bm{Q}\right)\right).
    % F_{enc} &= \mathcal{N}_{enc} \left(F\right) \\
    % F_{dec} &= \mathcal{N}_{dec} \left(F_{enc}, Q\right)\\
    \label{eq: detr}
\end{aligned}
\end{equation}
\begin{figure*}[!bt]
    \centering
    \subfigure[]{
    \includegraphics[height=5.5cm]{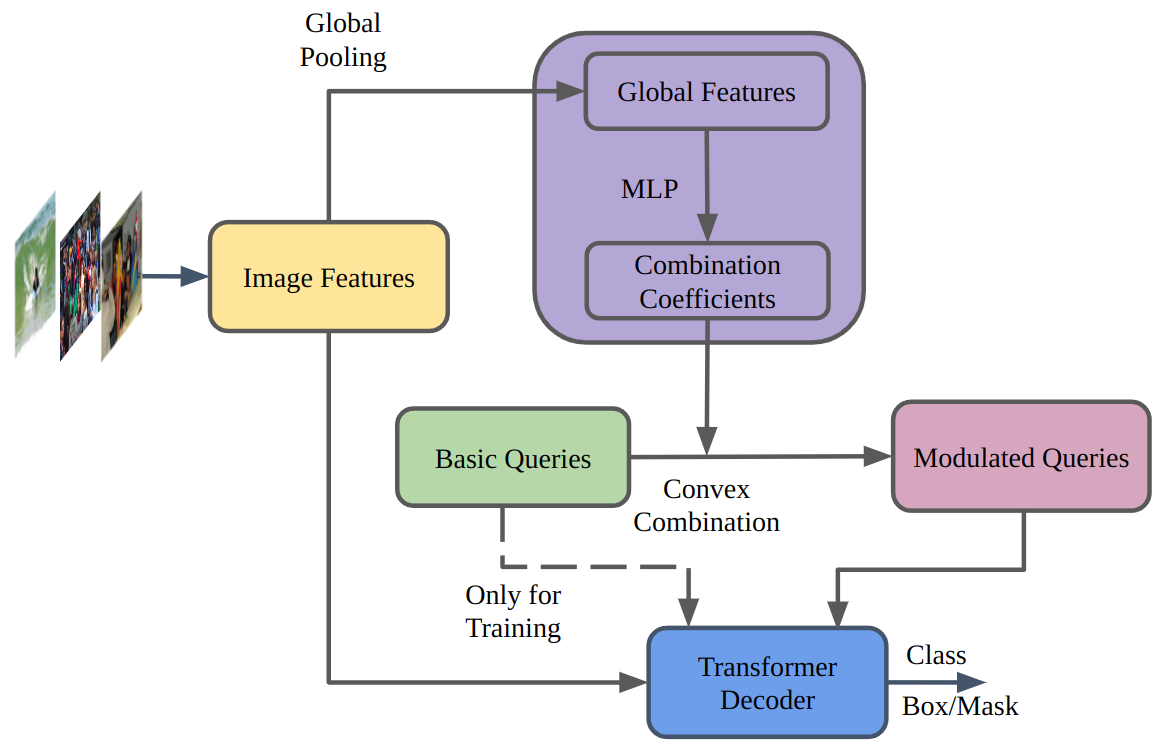}
    }
    \subfigure[]{
    \includegraphics[height=5.5cm]{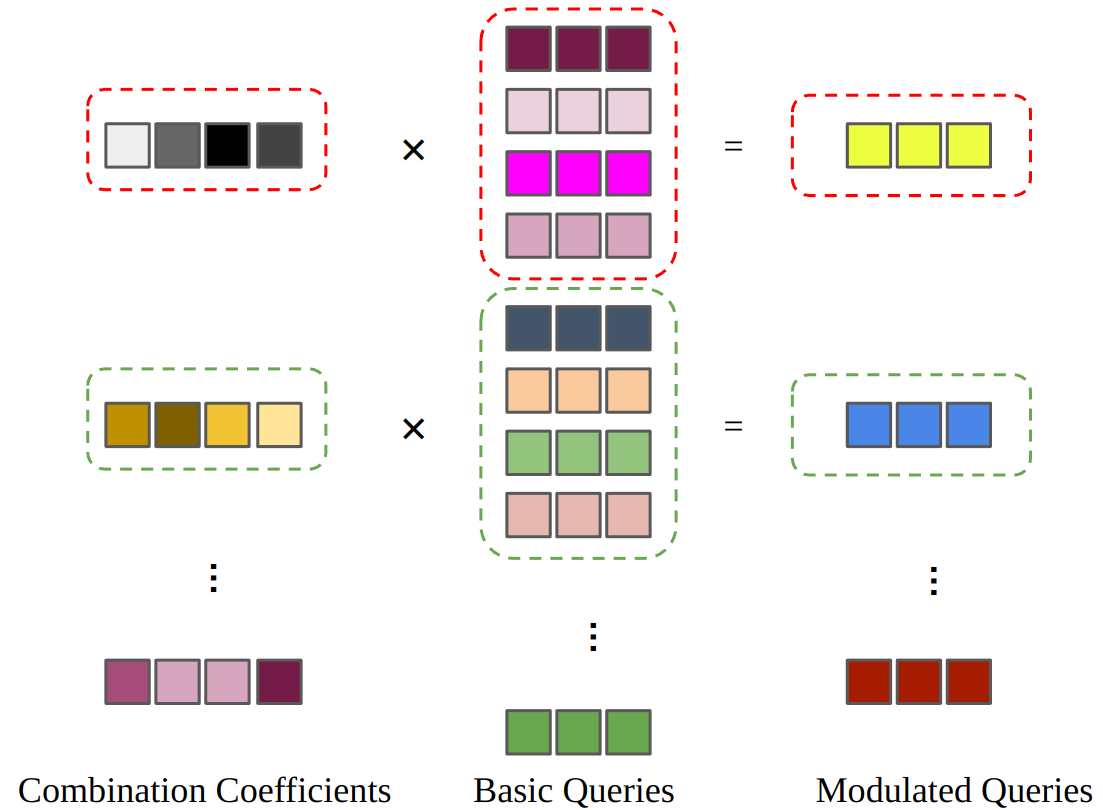}
    }
        \vspace{-0.3cm}
    \caption{The framework of the proposed method. (a) Model pipeline with dynamic query combinations. The step with the dashed line is only used in training. (b) Illustration of generating modulated queries from basic queries given combination coefficients.}
     \vspace{-0.4cm}
    \label{fig: dynamicQueryComvi}
\end{figure*}
For the object detection task, given the input image $\bm{I}$, multi-scale features $\bm{F}$ are extracted from the backbone network and then fed into a Transformer encoder $\mathcal{N}_{enc}$. After processing the features with multiple encoder layers, the output features are fed into a Transformer decoder $\mathcal{N}_{dec}$ together with $n$ randomly initialized query vectors $\bm{Q}\in\mathbb{R}^{n\times f}$, where $n$ and $f$ denote the number of queries and length of each query respectively. Each query can be a feature vector~\citep{detr,zhu2021deformable}, or a learned anchor box~\citep{liu2022dabdetr}. The outputs of $\mathcal{N}_{dec}$ are then fed into a task head $\mathcal{N}_{t}$ to generate the final predictions $\bm{Y} = \left\{\left(\bm{b}_i, \bm{c}_i\right), i = 1, 2, \dots, n\right\}$, 
where $\bm{b}_i, \bm{c}_i$ represent the bounding boxes and their corresponding categories of the detected objects. Then, the predictions are matched with the ground truths $\bm{Y}^\star$ using \cui{the} Hungarian Algorithm \citep{detr} to generate a bipartite matching. Then, the final loss is computed based on this bipartite matching: 
% \haichao{should we change $\mathcal{L}$ to $l$ since it is a scalar?} \cui{This is because that what is used in DETR paper for the loss of Hungarian loss.}
\begin{equation}
    \mathcal{L} = \mathcal{L}_\text{Hungarian}\left(\bm{Y}, \bm{Y}^\star\right).
\end{equation}
For the segmentation tasks, the final predictions are updated to $\bm{Y} = \left\{\left(\bm{b}_i, \bm{c}_i, \bm{m}_i\right), i=1,2,\dots,n\right\}$, where $\bm{m}_i$ denotes the predicted masks for different object instances. Since there is no direct correspondence of the predictions with the ground truth annotations, a bipartite matching is also computed to find the correspondence of the predictions and the ground truths $\bm{Y}^\star$. The final loss is then computed based on the matching. In some models such as Mask2Former \citep{cheng2021mask2former}, there will be no Transformer encoder $\mathcal{N}_{enc}$ to enhance the feature representations, while the other computational components follow the same paradigm. 
\subsection{Fixed Query Combinations}
% Our proposed method is an end-to-end framework with a backbone, an encoder-decoder module with multi-head self-attention and several prediction heads. 
% There are plenty of works proposed to improve the performance or convergence/inference speed of DETR \citep{detr} in recent years. 
Though some existing works analyze the contents of the queries for the decoder, such as Conditional-DETR \citep{tian2020conditional} and Anchor-DETR \citep{wang2022anchor}, they always exam each query individually. To the best of our knowledge, there is no work studying the interaction between the queries in $\bm{Q}$. Here, we would like to explore what kind of transformations conducted between the learned queries still generate ``good'' queries. If we compute the average of a few queries, is it still an effective query? If we use different types of linear transformations, which would be better to produce good queries?
% Also, among these existing methods, their performances are heavily influenced by the number of queries. By simply increasing the number of queries $N$, their performance can always be improved at the sacrifice of inference speeds or computational complexity. For example, by increasing the number of queries in Mask2Former \citep{cheng2021mask2former} from $50$ to $200$, the performance of instance and panoptic segmentation on MS COCO benchmarks \citep{lin2014microsoft} will be increased by $1.1\%$ on mAP and $1.7\%$ on PQ, respectively. However, most of the existing methods pay attention to how to improve the performance with well-designed queries rather than reducing the number of queries while preserving the performance for computational simplicity.
% To remove the gaps, we introduce two types of queries: basic queries $\bm{Q}^B  \in\mathbb{R}^{n\times f}$ and modulated queries $\bm{Q}^M \in \mathbb{R}^{m\times f}$, where $n = rm$ represent the number of queries, $r$ is set to be an integer larger than $1$ by default, and $f$ denotes the feature channels. Then, we conduct experiments to analyze the effects of generating a modulate query $q^m_i$ by combining multiple basic queries $q_j^b, j=1,2,\dots, r$ as the inputs of the Transformer decoder $\mathcal{N}_{dec}$, formulated as:
% where $w_j$ are the coefficients for the query combination process.

We conduct experiments to analyze the results of queries generated by different perturbations from the original queries. The procedure of the experiments is as follows: given a well-trained Transformer-based model, the initial queries for the decoder are denoted as $\bm{Q}^P=\left\{\bm{q}^P_1, \bm{q}^P_2, \dots, \bm{q}^P_n\right\} \in \mathbb{R}^{n\times f}$. The first type of perturbation \cui{uses} linear combinations of the original queries. We first separate the $n$ queries into $m$ groups, where each group has $r=\frac{n}{m}$ queries and generates one new query. Then, we initialize
% $r$ coefficients
\haichao{the combination coefficients}
$\bm{W}\in\mathbb{R}^{m\times r}$, where $w_{ij}\in\bm{W}$ is the coefficient used for the $i$-th group, $j$-th queries, denoted as $\bm{q}^P_{ij}$, to generate a group of new queries $\bm{Q}^{C} = \{\bm{q}^C_1, \bm{q}^C_2, \dots, \bm{q}^C_m\} \in\mathbb{R}^{m\times f}$. The process can be summarized as:  
\begin{equation}
    % \vspace{-0.1cm}
    \bm{q}^C_i = \sum_{j=1}^{r} w_{ij} \bm{q}^P_{ij} ,
    \vspace{-0.2cm}
    \label{eq: queryComb}
\end{equation}
We use three settings to evaluate the impact of different coefficients in Equation~\ref{eq: queryComb}, namely Convex Combination, Non-convex Combination, and Averaged Combination:

In Convex Combination, $\bm{q}^C_i$ is within the convex hull of $\bm{q}^P_{ij}, j=1,2,\dots,r$. The combination coefficients $w_{ij}$ are randomly initialized using uniform distribution in $\left[-1, 1\right]$ and then passed through a softmax function to satisfy the criteria: $w_{ij} \geq 0, \quad \sum_{j=1}^r w_{ij} = 1$.

For Non-convex Combinations, $w_{ij}$ are initialized in the same way as those in the convex combination, and the sum of $w_{ij}$ is forced to be $1$. However, there is no guarantee on its range and $w_{ij}$ can be negative values. For Averaged Combination, we generate $\bm{q}^C_i$ by averaging  $\bm{q}^P_{ij}, j=1,2,\dots,r$. As a baseline, we evaluate the model on $m$ queries randomly sampled from $\bm{Q}^P$. The experiments are conducted on MS COCO benchmark \citep{lin2014microsoft} for object detection, and instance segmentation, using DAB-DETR \citep{liu2022dabdetr}, Deformable-DETR~\citep{zhu2021deformable} and Mask2Former \citep{cheng2021mask2former}, with ResNet-50 \citep{he2016deep} as the backbone. The results are summarized in Table \ref{tab: convex}. 

From Table \ref{tab: convex}, we notice that Convex Combination achieves the best results among all the compared settings except the baseline. Convex Combination only degenerates slightly compared with learned queries on DAB-DETR and Mask2Former. 
In addition, the performance of Convex Combination only has very small variances across different models, proving that convex combinations of the group-wise learned queries are naturally high-quality object queries for different Transformer-based models on both detection and segmentation tasks. $n$ is set to 300 for detection models and 100 for Mask2Former. We run each experimental setting 6 times to compute the variance.
% first conduct experiments with fixed query combinations to analyze the influence of fusing multiple queries $\bm{Q}^P=\left\{\bm{q}^P_1, \bm{q}^P_2, \dots, \bm{q}^P_n\right\} \in\mathbb{R}^{n\times f}$ on the object detection performance with a pretrained model and use different strategies to generate multiple groups of coefficients $w_j$, which are then shared regardless of the input images. 

% For model A, we randomly initialize $w_j$ and feed them into a softmax function, so that $0 \leq w_j \leq 1$ and $\sum_{j=1}^r w_j = 1$. We follow the same pipeline as model A but multiply each coefficient $w_j$ with $-1$ to get the negative values to generate model B. For model C, we randomly generate the coefficients without any constraints. For comparison, we randomly initialize the queries without combining any basic queries to get model D and model E is the original model. For a better understanding, we use a pretrained DAB-DETR \citep{liu2022dabdetr} as an example. During the inference time, for model A, B and C, we replace the original queries in DAB-DETR with the query combination following Equation \ref{eq: queryComb}. For model D, we use randomly initialized queries instead of the original ones. Quantitative comparisons between these models are summarized ins Table \ref{tab: convex}.
\begin{table*}[!bt]
    \centering
    \begin{tabular}{l|l|l|l|l}
    \toprule
    \textcolor{black}{Backbone} & Method & mAP & AP$_\text{0.5}$ & AP$_\text{0.75}$\\
    \midrule
    \multirow{10}{*}{\rotatebox{0}{ResNet-50}} & Conditional-DETR \citep{tian2020conditional} & 40.9 & 61.7 & 43.3 \\
    & DQ-Conditional-DETR & 42.0$_{\uparrow 1.1}$ & 63.3$_{\uparrow1.6}$ & 44.2$_{\uparrow0.9}$ \\
    & SMCA-DETR \citep{gao2021fast} & 41.0 & 61.5 & 43.5 \\
    & DQ-SMCA-DETR& 42.1$_{\uparrow1.1}$ & 63.3$_{\uparrow1.8}$ & 44.9$_{\uparrow1.4}$ \\
    & DAB-DETR \citep{liu2022dabdetr} & 42.1 & 63.1 & 44.6\\
    & DQ-DAB-DETR& 43.7$_{\uparrow1.6}$ & 64.4$_{\uparrow1.3}$ & 46.6$_{\uparrow2.0}$ \\
    & Deformable-DETR \citep{zhu2021deformable} & 46.2 & 65.0 & 49.9 \\
    & DQ-Deformable-DETR  & 47.0$_{\uparrow0.8}$ & 65.5$_{\uparrow0.5}$ & 50.9$_{\uparrow1.0}$ \\
    & DAB-Deformable-DETR \citep{liu2022dabdetr} & 48.1 & 66.4 & 52.0 \\
    & DQ-DAB-Deformable-DETR & 49.7$_{\uparrow1.6}$ & 68.1$_{\uparrow1.7}$ & 54.2$_{\uparrow2.2}$ \\
    \midrule
    \multirow{4}{*}{\rotatebox{0}{Swin-Base}} & \textcolor{black}{Deformable-DETR \citep{zhu2021deformable}} &  50.9 & 70.5 & 55.3 \\
    & \textcolor{black}{DQ-Deformable-DETR} & 53.2$_{\uparrow2.3}$ & 72.8$_{\uparrow2.3}$ & 57.7$_{\uparrow2.4}$ \\
    & \textcolor{black}{DAB-Deformable-DETR \cite{liu2022dabdetr}} & 52.7 & 71.8 & 57.4\\
    & \textcolor{black}{DQ-DAB-Deformable-DETR} & 53.8$_{\uparrow1.1}$ & 72.8$_{\uparrow1.0}$ & 58.6$_{\uparrow1.2}$\\
    \bottomrule
    \end{tabular}
        \vspace{-0.3cm}
    \caption{Comparison of existing DETR-based object detectors with/without our proposed methods integrated on MS COCO \texttt{val} split.}
     \vspace{-0.4cm}
    \label{tab: objDet}
\end{table*}
\subsection{Dynamic Query Combinations}
From the previous section, we learn that fixed convex combinations of learned queries can still produce a reasonable accuracy compared to the learned queries. In this section, we propose a strategy to learn dynamic query combinations for the Transformer-based models instead of randomly generating the coefficients $w_{ij}$
for query combinations. Our model predicts their values according to the high-level content of the input. Thus, each input image will have a distinct set of object queries fed into the Transformer decoder.

\cui{To generate dynamic queries, a naive idea is to generate the queries directly from the input features $\bm{F}$. This method will increase the number of parameters dramatically, causing it difficult to optimize and inevitably computationally inefficient.
% Different from this, we borrow the idea from the dynamic convolution \citep{chen2020dynamic}, which aggregates the features with multiple kernels in each convolutional layer. Given a subset of basic queries, $q_j^b, j = 1, 2, \dots, r$, the corresponding modulated query $q_i^m$ is calculated as follows:
To verify this, we conduct an experiment on Deformable-DETR \citep{zhu2021deformable} with ResNet-50 as the backbone. We replace the original randomly initialized queries with those generated by a multi-layer perceptron (MLP), which transforms the image feature $\bm{F}$ to $\bm{Q}$. With 50 epochs, the model only achieves $45.1\%$ mAP, which is lower than the original model with $46.2\%$}.

\haichao{Inspired by the dynamic convolution \citep{chen2020dynamic, yang2019condconv}, which aggregates the features with multiple kernels in each convolutional layer, we propose a query modulation method.}
% Given a subset of basic queries, $q_j^b, j = 1, 2, \dots, r$, the corresponding modulated query $q_i^m$ is calculated as follows:
We introduce two types of queries: basic queries $\bm{Q}^B  \in\mathbb{R}^{n\times f}$ and modulated queries $\bm{Q}^M \in \mathbb{R}^{m\times f}$, where $n, m$ are the number of queries and $n = rm$. Equation \ref{eq: queryComb} is updated as:
\begin{equation}
    % \vspace{-0.1cm}
    \bm{q}^M_i = \sum_{j=1}^{r} w_{ij}^D \bm{q}^B_{ij},
    % \vspace{-0.2cm}
    \label{eq: dynamicQueryComb}
\end{equation}
where $\bm{W}^D \in \mathbb{R}^{m\times r}$ is the combination coefficient matrix and $w^D_{ij} \in \bm{W}^D$ is the coefficient for the $i$-th group, $j$-th query in $\bm{Q}^B$, denoted as $\bm{q}^B_{ij}$. To guarantee our query combinations to be convex, we add extra constraints to the coefficients as $w_{ij}^D \geq 0, \quad \sum_{j=1}^r w_{ij}^D=1$.

We use an example here to illustrate how to divide the basic queries into multiple groups. The basic queries are represented as $\bm{Q}^B = \{\bm{q}^B_0, \bm{q}^B_1,\bm{q}^B_2, \bm{q}^B_3,\bm{q}^B_4, \bm{q}^B_5,\bm{q}^B_6, \bm{q}^B_7\}$ and $r=4$. We divide the basic queries in sequential order. Therefore, $\bm{q}^B_0, \bm{q}^B_1,\bm{q}^B_2, \bm{q}^B_3$ is used to generate $\bm{q}^M_0$ and $\bm{q}^B_4, \bm{q}^B_5,\bm{q}^B_6, \bm{q}^B_7$ is used to generate $\bm{q}^M_1$. $\bm{w}_0^D \in\mathbb{R}^4$ is used to weighted average $\bm{q}^B_0, \bm{q}^B_1,\bm{q}^B_2, \bm{q}^B_3$ to generate $\bm{q}^M_0 \in \mathbb{R}^f$ while $\bm{w}_1^D \in\mathbb{R}^4$ is used to weighted average $\bm{q}^B_4, \bm{q}^B_5,\bm{q}^B_6, \bm{q}^B_7$ to generate $\bm{q}^M_1 \in \mathbb{R}^f$. We did not conduct experiments to study the effects of different divisions. Since the basic queries are randomly initialized and are jointly learned with the modulated queries, we believe the results will not change significantly with a different division.

In our dynamic query combination module, the coefficient matrix $\bm{W}^{D}$ is learned based on the input feature $\bm{F}$ through a mini-network, as:
\begin{equation}
    \bm{W^D} = \sigma\left(\theta\left(\mathcal{A}\left(\bm{F}\right)\right)\right),
    \label{eq: constraints}
    % \vspace{-0.4cm}
\end{equation}
where $\mathcal{A}$ is a global average pooling to generate a global feature from the feature map $\bm{F}$, $\theta$ is an MLP, $\sigma$ is a softmax function to guarantee the elements of $\bm{W}^D$ satisfy the convex constraints. \textcolor{black}{Here we try to make the mini-network as simple as possible to show the potential of using modulated queries. This attention-style structure happens to be a simple and effective design choice.  }

During the training process, we feed both $\bm{Q}^M$ and $\bm{Q}^B$ to the same decoder to generate the corresponding predictions $\bm{Y}^M$ and $\bm{Y}^B$ as follows, 
\begin{equation}
\begin{aligned}
    \bm{Y}^{M} &= \mathcal{N}_{t} \left(\mathcal{N}_{dec}\left(\mathcal{N}_{enc}\left(\bm{F}\right), \bm{Q}^M\right)\right)\\
    \bm{Y}^B &= \mathcal{N}_{t} \left(\mathcal{N}_{dec}\left(\mathcal{N}_{enc}\left(\bm{F}\right), \bm{Q}^B\right)\right)\\
    \label{eq: bothQuery}
\end{aligned}
    \vspace{-0.5cm}
\end{equation}
The final training loss is then updated to 
\begin{equation}
\begin{aligned}
    \mathcal{L} &= \mathcal{L}_\text{Hungarian}\left(\bm{Y}^M, \bm{Y}^{\star}\right) + \beta\mathcal{L}_\text{Hungarian}\left(\bm{Y}^B, \bm{Y}^{\star}\right)   \label{eq: newLoss}
\end{aligned}
\end{equation}
where $\beta$ is a hyperparameter.
During the inference, only $\bm{Q}^M$ is used to generate the final predictions $\bm{Y}^M$ while the basic queries $\bm{Q}^B$ are not used. Therefore, the computational complexity increases for our models are negligible compared to the original DETR-based models. The only difference in the computation is that we have an additional MLP and a convex combination to generate the modulated queries. Therefore, the role of modulated queries in our model is exactly the same as the fixed object queries in the original models.
% we use \emph{modulated queries} and \emph{queries} interchangeably to refer to the modulated queries.
\begin{table*}[!bt]
    \centering
    \begin{tabular}{l|l|l|l|l|l|l}
    \toprule
     \multirow{2}{*}{Method} & \multirow{2}{*}{Backbone} & \multicolumn{3}{c|}{Panoptic} & \multicolumn{2}{c}{Instance}\\
     \cline{3-7}
     \rule{0pt}{12pt} 
     & & PQ  & $\text{AP}_{\text{pan}}^{\text{Th}}$ & mIoU$_\text{pan}$ & mAP & AP$_{0.5}$\\
    \midrule
    Mask2Former \citep{cheng2021mask2former} & \multirow{2}{*}{ResNet-50} & 62.1 & 37.3 & 77.5 & 37.4 & 61.9\\
    DQ-Mask2Former & & 63.2$_{\uparrow1.1}$ & 38.2$_{\uparrow0.9}$ & 78.7$_{\uparrow1.2}$ & 38.5$_{\uparrow1.1}$  & 63.2$_{\uparrow1.3}$ \\
    \midrule
    Mask2Former \citep{cheng2021mask2former} & \multirow{2}{*}{Swin-Base} & 66.1 & 42.8 & 82.7 & 42.0 & 68.8\\
    DQ-Mask2Former & & 67.0$_{\uparrow0.9}$ & 43.7$_{\uparrow0.9}$ & 83.7$_{\uparrow1.0}$ & 43.0$_{\uparrow1.0}$ & 69.6$_{\uparrow0.8}$ \\
    \bottomrule
    \end{tabular}
        \vspace{-0.3cm}
    \caption{Comparison of Mask2Former and DQ-Mask2Former on panoptic and instance segmentation tasks on CityScapes \texttt{val} split.}
     \vspace{-0.4cm}
    \label{tab: cityscape}
\end{table*}

\begin{table}[!bt]
    \centering
    \footnotesize
    \begin{tabular}{l|l|l}
    \toprule
    Methods & Backbone & mAP\\
    \midrule
    Mask R-CNN \citep{he2017mask} & \multirow{2}{*}{ResNet-50} & 35.4\\
    QueryInst \citep{Fang_2021_ICCV} & & 39.8\\
    \midrule
    \textcolor{black}{Mask2Former\citep{cheng2021mask2former}} & \multirow{2}{*}{ResNet-50}  & 43.7\\
    \textcolor{black}{DQ-Mask2Former} & & 44.4$_{\uparrow0.7}$\\
    \midrule
    \textcolor{black}{Mask2Former \citep{cheng2021mask2former}} & \multirow{2}{*}{Swin-Base}  & 46.7\\
    \textcolor{black}{DQ-Mask2Former} & & 47.6$_{\uparrow0.9}$ \\
    \bottomrule
    \end{tabular}
        \vspace{-0.3cm}
    \caption{Comparison of existing instance segmentation approaches and DQ-Mask2Former on MS COCO \texttt{val} split.}
     \vspace{-0.4cm}
    \label{tab: insCoco}
\end{table}
\section{Experiments}
To evaluate the effectiveness of our proposed methods, we first conduct experiments on a series of tasks, including object detection, instance segmentation, panoptic segmentation, and video instance segmentation with different DETR-based models. Then we conduct several ablation studies to investigate the impact of different hyperparameters in our model for a better analysis. Finally, we visualize the dynamic query combinations to show the effectiveness of our model.
\vspace{-0.1cm}
\subsection{Experiment Setup}
\noindent\textbf{Datasets.} For the object detection task, we use MS COCO benchmark \citep{lin2014microsoft} for evaluation, which contains $118,287$ images for training and $5,000$ for validation. For instance and panoptic segmentation, besides the MS COCO benchmark ($80$ ``things" and $53$ ``stuff" categories), we also conduct experiments on the CityScapes \citep{Cordts2016Cityscapes} benchmark ($8$ ``things" and $11$ ``stuff" categories) to validate the effectiveness of our proposed method. For the video instance segmentation task, YouTube-VIS-2019 \citep{yang2019video} is used for evaluation. For experiments on video instance segmentation, we pretrain our models on MS COCO and finetune them on the training set of YouTube-VIS-2019. 
% In the manuscript, we only provide experiment results on MS COCO and YouTube-VIS-2019 benchmarks.  
%Following the pipelines of Deformable-DETR \citep{zhu2021deformable} and Mask2Former \citep{cheng2021mask2former}, our models are trained on the training set and then evaluated on the validation set.

\noindent\textbf{Evaluation metrics.} For panoptic segmentation, the standard \textbf{PQ} (panoptic quality) metric \citep{kirillov2019panoptic} is used for evaluation. 
% Meanwhile, we report the \textbf{ AP$^\text{Th}_\text{pan}$} and \textbf{$\text{mIOU}_\text{pan}$} as Mask2Former \citep{cheng2021mask2former}.
For instance segmentation (image or video) and object detection, we use the standard \textbf{mAP} (mean average precision) metric for evaluation. For VIS, mAP and AR (average recall) on video instances are the evaluation metrics. We observe around $0.8$ mAP fluctuations in performance and we report  the results reproduced based on the officially released code in this section.

\noindent\textbf{Implementation details.} The query ratio $r$ used to generate the combination coefficients is set to $4$ by default. $\beta$ is set to be $1$. $\theta$ is implemented as a two-layer MLP with ReLU as nonlinear activations. The output size of its first layer is 512, and that of the second layer is the length of $\bm{W}^D$ in corresponding models. For detection models, we use 300 modulated queries and 1200 basic queries if not specified otherwise. For the baseline models used for comparison, we use 300 queries as the original implementation by default. For instance segmentation and panoptic segmentation models, we use 100 modulated queries and 400 basic queries for Mask2Former. For video instance segmentation, we use 100 modulated queries and 400 basic queries for Mask2Former and 300 modulated queries, and 1,200 basic queries for SeqFormer. For the baseline models, we use 100 queries for Mask2Former on image segmentation tasks, 100 queries for Mask2Former on video instance segmentation, and 300 queries for SeqFormer on video instance segmentation. The comparison is based on the fairness principle that the same number of queries are used as input to the transformer decoders of our model and the baseline. During inference, the only added computational cost of our method compared to the baselines is the mini-network to produce the modulated queries. Code is available at \url{https://github.com/bytedance/DQ-Det}.

\begin{table*}[!bt]
    \centering
    \begin{tabular}{l|l|l|l}
    \toprule
    Methods & Backbone &  PQ & PQ$_\text{th}$ \\
    \midrule
    UPSnet \citep{xiong2019upsnet} & \multirow{2}{*}{ResNet-50} & 42.5 & 48.6\\
    DETR \citep{detr} & & 43.4 & 48.2\\
    \midrule
    Mask2Former \citep{cheng2021mask2former} & \multirow{2}{*}{ResNet-50} & 51.9 & 57.7\\
    DQ-Mask2Former & & 52.6$_{\uparrow0.7}$ & 58.9$_{\uparrow1.2}$\\
    \midrule
    Mask2Former \citep{cheng2021mask2former} & \multirow{2}{*}{Swin-Base} & 55.1 & 61.0\\
    DQ-Mask2Former & & 55.7$_{\uparrow0.6}$ & 61.8$_{\uparrow0.8}$\\
    \bottomrule
    \end{tabular}
        \vspace{-0.3cm}
    \caption{Comparison of existing panoptic segmentation approaches with DQ-Mask2Former on MS COCO \texttt{val} split.}
     \vspace{-0.4cm}
    \label{tab: panoCoco}
\end{table*}
\subsection{Main Results}
\noindent\textbf{Object detection.}  We evaluate our proposed methods with the DETR-based models Deformable-DETR \citep{zhu2021deformable}, SMCA-DETR \citep{gao2021fast}, Conditional-DETR \citep{tian2020conditional}, DAB-DETR and DAB-Deformable-DETR \citep{liu2022dabdetr} with ResNet50~\cite{he2016deep} for object detection on the MS COCO benchmark. We also experiment with  Deformable-DETR and DAB-Deformable-DETR on Swin-B\cite{liu2021swin} to further evaluate the performance of our method on more powerful backbones. For a fair comparison, we run the original model integrated with and without our proposed modulated queries using the same experimental settings. The models equipped with our dynamic query combinations are denoted as DQ-Deformable-DETR, DQ-SMCA-DETR, DQ-Conditional-DETR,  DQ-DAB-DETR, and DQ-DAB-Deformable-DETR, respectively. The results are shown in Table \ref{tab: objDet}. From Table \ref{tab: objDet}, when integrated with our proposed method, mAP can be improved consistently by at least $0.8\%$ for all the models listed in the table. \textcolor{black} {For DAB-Deformable-DETR, the mAP can be improved by $1.6\%$ with ResNet50 backbone and $1.1\%$ with Swin-Base backbone. For Deformable-DETR, the mAP can be improved significantly by $2.3\%$ with the Swin-Base backbone. This proves the benefit of our method with different backbones.} Note that models with modulated queries only have negligible increased computation cost compared to the original models.

\noindent\textbf{Instance/panoptic segmentation.} Mask2Former\citep{cheng2021mask2former} is a recent state-of-the-art model that can be used for different segmentation tasks with a unified model architecture. We compare Mask2Former  with/without our modulated queries for image instance and panoptic segmentation tasks on the MS COCO \citep{lin2014microsoft} and CityScapes benchmarks. The model plugged with modulated queries is named DQ-Mask2Former. The results are shown in Table \ref{tab: cityscape}, Table \ref{tab: insCoco}, and Table \ref{tab: panoCoco}, respectively. 

For instance segmentation (Table \ref{tab: insCoco}), our model DQ-Mask2Former achieves consistent improvement across different metrics compared to the original Mask2Former. For example, the performance on mAP is improved by around $0.8\%$ on both MS COCO. For panoptic segmentation, as shown in Table \ref{tab: panoCoco}, DQ-Mask2Former again significantly outperforms Mask2Former \citep{cheng2021mask2former} across all the evaluation metrics on both MS COCO and CityScapes. Since panoptic segmentation is more challenging compared with instance segmentation and object detection where both semantic and instance segmentation tasks are required to generate the final predictions, our model works less effectively for panoptic segmentation compared to other tasks.

\noindent\textbf{Video instance segmentation.} Besides image tasks, we also evaluate our method on the video instance segmentation task. We evaluated our method based on two state-of-the-art video instance segmentation methods Mask2Former~\citep{cheng2021mask2former} and SeqFormer~\citep{wu2022seqformer}. Results are shown in Table \ref{tab: vis}. It can be seen from Table \ref{tab: vis} that when integrated with our modulated queries, mAP, and AR of Mask2Former are improved by at around $1.0\%$ and the mAP of SeqFormer is significantly boosted by $1.5\%$. Note the additional computation cost is negligible  with our method.
\begin{table*}[!bt]
    \centering
     \begin{tabular}{l|l|l|l|l}
    \toprule
     Method  & Backbone & mAP & AP$_\text{0.5}$ & AP$_\text{0.75}$ \\
    \midrule
    MaskTrack R-CNN \citep{yang2019video} & \multirow{2}{*}{ResNet-50} & 30.3 & 51.1 & 32.6\\
    IFC \citep{hwang2021video} & & 42.8 & 65.8 & 46.8\\
    \midrule
    Mask2Former \citep{cheng2021mask2former} & \multirow{4}{*}{ResNet-50} & 46.4 & 68.0 & 50.0\\
    DQ-Mask2Former & & 47.4$_{\uparrow1.0}$ & 69.2$_{\uparrow1.2}$ & 51.0$_{\uparrow1.0}$ \\
    SeqFormer \citep{wu2022seqformer} & & 47.4 & 69.8& 51.8\\
    DQ-SeqFormer & & 49.0$_{\uparrow1.6}$ & 71.5$_{\uparrow1.7}$ & 53.0$_{\uparrow1.2}$\\
    \midrule
    Mask2Former \citep{cheng2021mask2former} & \multirow{2}{*}{Swin-Base} & 59.5 & 84.3 & 67.2\\
    DQ-Mask2Former & & 61.3$_{\uparrow1.8}$ & 86.1$_{\uparrow1.8}$ & 68.6$_{\uparrow1.4}$ \\
    \midrule
    SeqFormer \citep{wu2022seqformer} & \multirow{2}{*}{Swin-Large} & 59.3 & 82.1	& 66.4\\
    DQ-SeqFormer & & 61.2$_{\uparrow1.9}$ & 84.1$_{\uparrow2.0}$ & 68.0$_{\uparrow1.6}$\\
    \bottomrule
    \end{tabular}
    \vspace{-0.3cm}
    \caption{Comparison of existing video instance segmentation approaches with DQ-Mask2Former and DQ-SeqFormer on YouTube-VIS-2019 \texttt{val} split.}
     \vspace{-0.4cm}
    \label{tab: vis}
\end{table*}
\subsection{Model Analysis}
In this section, we conduct extensive experiments to analyze the designs of our proposed method. By default, for the object detection task, the number of modulated queries is set to $300$. For the segmentation tasks, the number of modulated queries is set to $50$ for a faster training pipeline, and $r$ is set to $4$ for all the tasks.

\noindent\textbf{Analysis of the number of queries.} We use Deformable-DETR and DAB-Deformable-DETR as baseline models to study the effects of the number of queries on the performance of object detection. We compare the baseline models with DQ-Deformable-DETR and DQ-DAB-Deformable-DETR integrated with different numbers of queries as in Figure \ref{fig: resultCompare}. Note that we include the additional components of our models in the FLOPs computation of the decoder. When integrated with our method, even by reducing the number of modulated queries from $300$ to $100$, the mAPs of DQ-Deformable-DETR and DQ-DAB-Deformable-DETR are still better than the baseline models with $300$ queries. We are also able to reduce the computation cost of the decoders of Deformable-DETR and DAB-Deformable-DETR by about $14\%$ and $24\%$ by using our method with 100 queries, respectively. However, we do not observe significant speedup using our method with fewer queries mainly because the main computation costs are from the backbones and the transformer encoders.
%Meanwhile, we notice that the number of queries does not have too much effect on the performance when integrated with our proposed module.
\begin{table}[!bt]
    \centering
        % \footnotesize
    \begin{tabular}{l|l|l|l|l|l|l}
    \toprule
    $\beta$ & mAP & AP$_\text{0.5}$ & AP$_\text{0.75}$ & AP$_\text{S}$ & AP$_\text{M}$ & AP$_\text{L}$ \\
    \midrule
    0.0 & 45.6 & 64.1 & 49.4 & 27.2 & 49.1 & 60.5\\
    0.5 & 46.4 & 65.0 & 50.3 & 28.1 & 49.2 & 62.6\\
    1.0 & 47.0 & 65.5 & 50.9 & 28.8 & 50.1 & 62.2\\
    \bottomrule
    \end{tabular}
    \vspace{-0.3cm}
    \caption{Analysis of $\beta$ using DQ-Deformable-DETR with ResNet-50 backbone on the MS COCO benchmark with different settings.}
     \vspace{-0.4cm}
    \label{tab: beta}
\end{table}

\begin{figure*}[!bt]
    \centering
    \subfigure[]{
    \includegraphics[height=5cm]{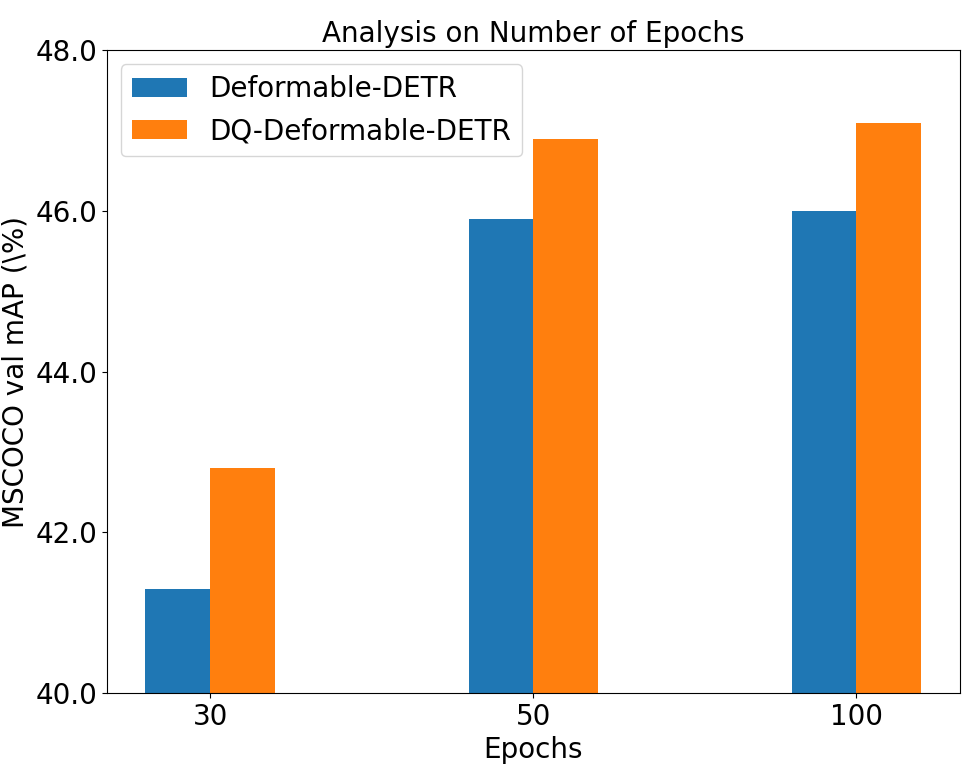}
    }
    \subfigure[]{
    \includegraphics[height=5cm]{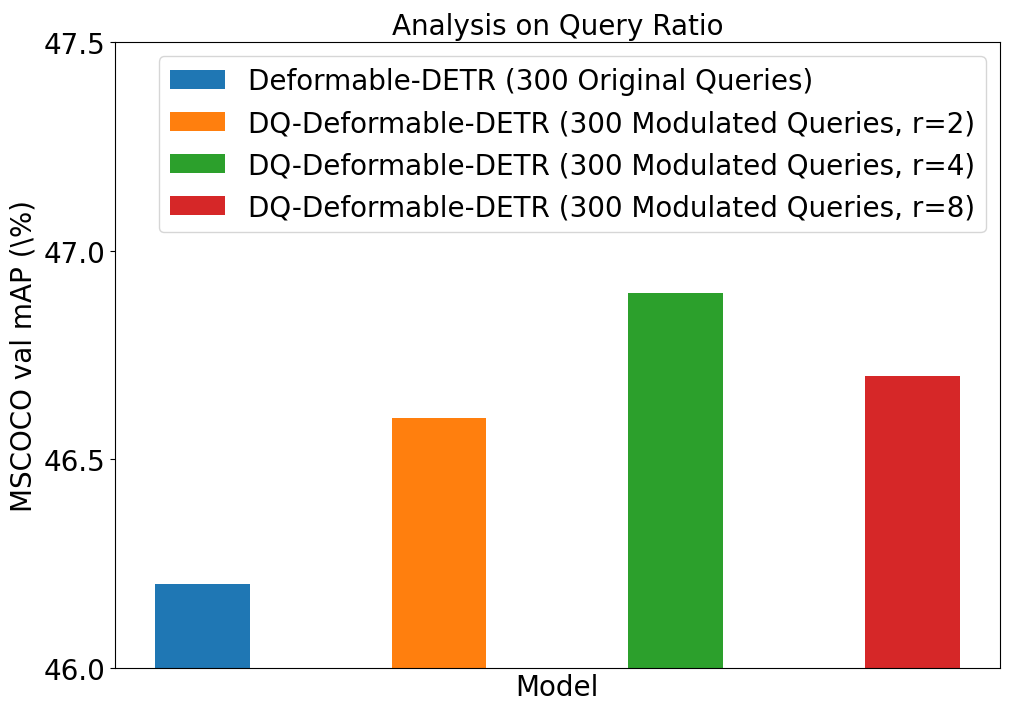}
    }
    \vspace{-0.4cm}
    \caption{Analysis of the impact of the number of epochs and query ratios on the performance.}
     \vspace{-0.4cm}
    \label{fig: analysis}
\end{figure*}

\noindent\textbf{Analysis of the number of training epochs.} In Figure~\ref{fig: analysis} (a), we show the impact of the number of training epochs on a sample model DQ-Deformable-DETR \cui{together with the original Deformable-DETR}.
\cui{From the figure, the mAP of DQ-Deformable-DETR is always better than that of the original Deformable-DETR at different epochs on the MS COCO benchmark. At early 30 epochs, DQ-Deformable-DETR achieves \cui{a more significant} performance gain compared to Deformable-DETR compared with later epochs.}

\noindent\textbf{Analysis of $\beta$.} \cui{We analyze the impact of the scale of $\beta$ on models equipped with our modulated queries. We conduct experiments using Deformable-DETR with ResNet-50 as the backbone of the MS COCO benchmark with different values of $\beta$. Results are shown in Table \ref{tab: beta}. As shown in the table, when $\beta$ is set to be $0$, where no loss is directly computed with the prediction from the basic queries, the performance drops by $2.4\%$ compared to the original setting. In this case, the basic queries are not necessarily proper queries for the detection model, which will affect the quality of the modulated queries produced by them. The performance can be improved by increasing the value of $\beta$. Empirically we find $\beta=1$ is a good choice to balance the scale of losses between the basic and modulated queries.}

\noindent\textbf{Analysis of query ratio.} We use DQ-Deformable-DETR \citep{zhu2021deformable} to analyze the performance of our proposed methods with different query ratios 2, 4, and 8, as in Figure \ref{fig: analysis} (b). From the figure, using $4$ as the query ratio achieves the best performance for DQ-Deformable-DETR with $300$ modulated queries. However, other query ratio choices still generate better accuracies than the original Deformable-DETR, which validates the effectiveness and robustness of our method.

\noindent\textbf{Non-modulated ablation.} Our model uses additional training queries (the basic queries) compared to the baselines. Here we conduct an ablative study on the factor of additional training queries. We train the model with two unrelated groups of queries with Deformable-DETR (1200/300 queries) and Mask2Former (400/100 queries), as Table \ref{tab: unrelated}. The first group has the same number as our basic queries while the second group has the same number  as our modulated queries. Only the second group is used in inference. From the table, models with the two unrelated groups produce similar results as the baselines. In contrast, our proposed method with modulated queries achieves significant improvement in the two models. This proves that the improvement of our model is not simply due to more queries in training on DETR-based models.

\begin{table}[!bt]
    \centering
        % \footnotesize
    \begin{tabular}{l|l|l|l}
    \toprule
    Method	& mAP	& AP$_\text{50}$	& AP$_\text{75}$\\
    \midrule 
    Deformable-DETR	& 46.2	& 65.0	&49.9 \\
    Deformable-DETR$^\dagger$	&46.5	&64.9	&50.5 \\ 
    DQ-Deformable-DETR	& 47.0 	&65.5	&50.9 \\
    \midrule
    Mask2Former	& 43.7	& 65.5	& 46.9 \\
    Mask2Former$^\dagger$	& 43.9 & 	65.8	&47.2 \\
    DQ-Mask2Former	& 44.4	&66.3	&47.6 \\
\bottomrule
\end{tabular}
\vspace{-0.3cm}
\caption{Comparison with two groups of unrelated queries. $\dagger$ denotes two groups of unrelated queries.}
 \vspace{-0.3cm}
\label{tab: unrelated}
\end{table}
\noindent\textbf{Analysis of speed.}
The proposed method needs to apply forward passes in two branches for the basic queries and the modulated queries in the training phase of the model, which increases the computation cost. Here, we report the  training time and inference speed of our model compared to the baseline with Deformable-DETR (ResNet-50) and Mask2Former (ResNet-50) in Table~\ref{tab: time}. The training time is based on 8 NVIDIA A100 GPUs and the inference FPS is tested on a single TITAN RTX GPU. From the table, our method only increases the training time slightly compared to the baselines. The reason is that the major computation cost of DETR-based models comes from the backbones and transformer encoders that only need to be forwarded once for the two branches. During inference, the FPS of DQ-Deformable-DETR and DQ-Mask2Former are slightly reduced by less than $4\%$ due to the extra computation of the mini-network to produce the modulated queries. 

\begin{table}[!bt]
    \centering
    \footnotesize
    \begin{tabular}{l|l|c}
    \toprule
Model	& Training Time & 	Inference FPS \\
\midrule
Deformable-DETR	& $\sim$ 61 GPU hours	& 13.3\\
DQ-Deformable-DETR	&  $\sim$ 69 GPU hours &	13.1 \\
\cline{1-3}
\rule{0pt}{10pt}
Mask2Former	&  $\sim$ 71 GPU hours	 & 5.4\\
DQ-Mask2Former	& $\sim$ 80 GPU hours	& 5.2\\
\bottomrule
\end{tabular}
\vspace{-0.5cm}
\caption{Comparison of the training/inference time with and without our proposed methods integrated.}
 \vspace{-0.4cm}
\label{tab: time}
\end{table}

\begin{figure}[!bt]
    \centering
    \includegraphics[width=8cm]{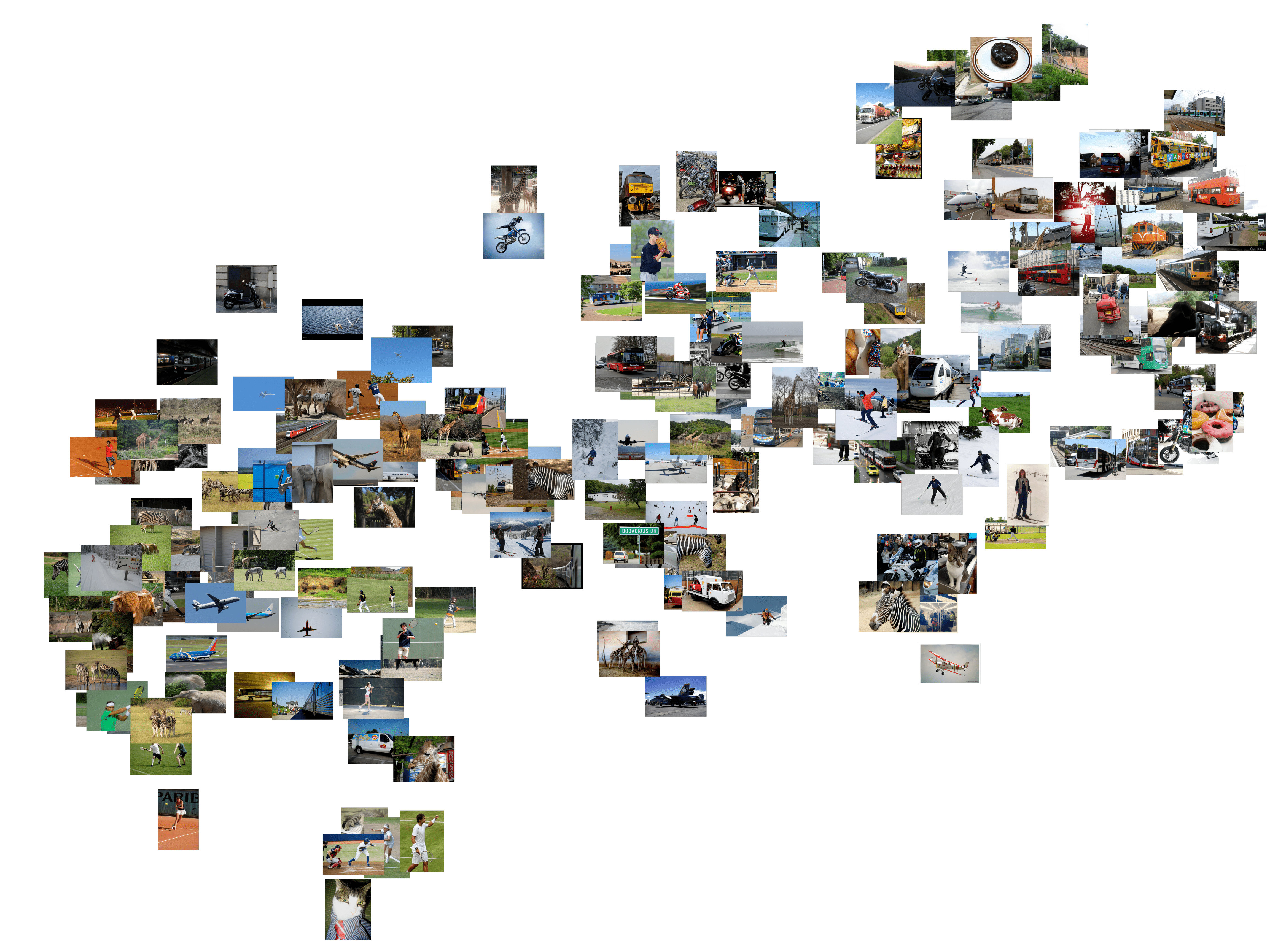}
        \vspace{-0.4cm}
    \caption{t-SNE visualization of $\bm{W}^D$ on 200 images from MS COCO \texttt{val}. Zoom in to see details.}
    \label{fig: tsne}
     \vspace{-0.6cm}
\end{figure}

\noindent\textbf{Visualization of $\bm{W}^D$.} Since $\bm{W}^D$ is conditioned on the high-level content of the image, we conjecture that images with similar scenes or object categories may have similar $\bm{W}^D$ parameters. We choose 200 images from the validation set of MS COCO and compute their $\bm{W}^D$ from DQ-DAB-DETR with 300 queries. The resulting $\bm{W}^D$ are first flattened into vectors and then projected onto a two-dimensional space using t-SNE \citep{van2008visualizing}. We visualize the projected $\bm{W}^D$ parameters along with their corresponding input images as Figure \ref{fig: tsne}. We can see that some object categories tend to be clustered. For example, we can see a lot of transportation vehicles in the top right corner of the figure, and wild animals tend to be in the lower part of the figure, which indicates that the model uses some high-level semantics of the image to produce the combination coefficients. 

\vspace{-0.15cm}
\section{Conclusion}
In this paper, we propose to use dynamic queries depending on the input image to enhance DETR-based detection and segmentation models. We find that convex combinations of learned queries are naturally high-quality object queries for the corresponding models. Based on this observation, we design a pipeline to learn dynamic convex combinations of the basic queries, adapting object queries according to the high-level semantics of the input images. This approach consistently improves the performance of a wide range of DETR-based models on object detection and segmentation tasks. The gain of our model is agnostic to the different designs of the Transformer decoders and different types of object queries. We believe this approach opens the door to designing dynamic queries and creates a new perspective for Transformer-based models.

% In the unusual situation where you want a paper to appear in the
% references without citing it in the main text, use \nocite
% \nocite{langley00}

\bibliography{ref}

\begin{thebibliography}{65}
\providecommand{\natexlab}[1]{#1}
\providecommand{\url}[1]{\texttt{#1}}
\expandafter\ifx\csname urlstyle\endcsname\relax
  \providecommand{\doi}[1]{doi: #1}\else
  \providecommand{\doi}{doi: \begingroup \urlstyle{rm}\Url}\fi

\bibitem[Bolya et~al.(2019)Bolya, Zhou, Xiao, and Lee]{yolact-iccv2019}
Bolya, D., Zhou, C., Xiao, F., and Lee, Y.~J.
\newblock Yolact: {Real-time} instance segmentation.
\newblock In \emph{ICCV}, 2019.

\bibitem[Bolya et~al.(2020)Bolya, Zhou, Xiao, and Lee]{yolact-plus-tpami2020}
Bolya, D., Zhou, C., Xiao, F., and Lee, Y.~J.
\newblock Yolact++: Better real-time instance segmentation.
\newblock \emph{IEEE Transactions on Pattern Analysis and Machine
  Intelligence}, 2020.

\bibitem[Cai \& Vasconcelos(2019)Cai and Vasconcelos]{Cai_2019}
Cai, Z. and Vasconcelos, N.
\newblock Cascade r-cnn: High quality object detection and instance
  segmentation.
\newblock \emph{IEEE Transactions on Pattern Analysis and Machine
  Intelligence}, pp.\  1–1, 2019.
\newblock ISSN 1939-3539.
\newblock \doi{10.1109/tpami.2019.2956516}.
\newblock URL \url{http://dx.doi.org/10.1109/tpami.2019.2956516}.

\bibitem[Cao et~al.(2020)Cao, Anwer, Cholakkal, Khan, Pang, and
  Shao]{cao2020sipmask}
Cao, J., Anwer, R.~M., Cholakkal, H., Khan, F.~S., Pang, Y., and Shao, L.
\newblock Sipmask: Spatial information preservation for fast image and video
  instance segmentation.
\newblock In \emph{ECCV}, 2020.

\bibitem[Cao et~al.(2019)Cao, Xu, Lin, Wei, and Hu]{cao2019gcnet}
Cao, Y., Xu, J., Lin, S., Wei, F., and Hu, H.
\newblock Gcnet: Non-local networks meet squeeze-excitation networks and
  beyond.
\newblock In \emph{ICCVW}, 2019.

\bibitem[Carion et~al.(2020)Carion, Massa, Synnaeve, Usunier, Kirillov, and
  Zagoruyko]{detr}
Carion, N., Massa, F., Synnaeve, G., Usunier, N., Kirillov, A., and Zagoruyko,
  S.
\newblock End-to-end object detection with transformers.
\newblock In \emph{ECCV}, 2020.

\bibitem[Chen et~al.(2022)Chen, Chen, Zeng, and Wang]{chen2022group}
Chen, Q., Chen, X., Zeng, G., and Wang, J.
\newblock Group detr: Fast training convergence with decoupled one-to-many
  label assignment.
\newblock \emph{arXiv preprint arXiv:2207.13085}, 2022.

\bibitem[Chen et~al.(2020)Chen, Dai, Liu, Chen, Yuan, and Liu]{chen2020dynamic}
Chen, Y., Dai, X., Liu, M., Chen, D., Yuan, L., and Liu, Z.
\newblock Dynamic convolution: Attention over convolution kernels.
\newblock In \emph{CVPR}, 2020.

\bibitem[Cheng et~al.(2021{\natexlab{a}})Cheng, Choudhuri, Misra, Kirillov,
  Girdhar, and Schwing]{cheng2021mask2former}
Cheng, B., Choudhuri, A., Misra, I., Kirillov, A., Girdhar, R., and Schwing,
  A.~G.
\newblock Mask2former for video instance segmentation.
\newblock \emph{arXiv preprint arXiv:2112.10764}, 2021{\natexlab{a}}.

\bibitem[Cheng et~al.(2021{\natexlab{b}})Cheng, Schwing, and
  Kirillov]{cheng2021per}
Cheng, B., Schwing, A., and Kirillov, A.
\newblock Per-pixel classification is not all you need for semantic
  segmentation.
\newblock \emph{NIPS}, 2021{\natexlab{b}}.

\bibitem[Cordts et~al.(2016)Cordts, Omran, Ramos, Rehfeld, Enzweiler, Benenson,
  Franke, Roth, and Schiele]{Cordts2016Cityscapes}
Cordts, M., Omran, M., Ramos, S., Rehfeld, T., Enzweiler, M., Benenson, R.,
  Franke, U., Roth, S., and Schiele, B.
\newblock The cityscapes dataset for semantic urban scene understanding.
\newblock In \emph{CVPR}, 2016.

\bibitem[Cui(2022)]{cui2022dynamic2}
Cui, Y.
\newblock Dynamic feature aggregation for efficient video object detection.
\newblock In \emph{ACCV}, 2022.

\bibitem[Cui(2023)]{Cui_2023_CVPR}
Cui, Y.
\newblock Feature aggregated queries for transformer-based video object
  detectors.
\newblock In \emph{CVPR}, 2023.

\bibitem[Cui et~al.(2022)Cui, Yang, and Liu]{cui2022dynamic}
Cui, Y., Yang, L., and Liu, D.
\newblock Dynamic proposals for efficient object detection.
\newblock \emph{arXiv preprint arXiv:2207.05252}, 2022.

\bibitem[Dong et~al.(2021)Dong, Zeng, Wang, Zhang, and Wei]{dong2021solq}
Dong, B., Zeng, F., Wang, T., Zhang, X., and Wei, Y.
\newblock Solq: Segmenting objects by learning queries.
\newblock \emph{NIPS}, 2021.

\bibitem[Duan et~al.(2019)Duan, Bai, Xie, Qi, Huang, and
  Tian]{duan2019centernet}
Duan, K., Bai, S., Xie, L., Qi, H., Huang, Q., and Tian, Q.
\newblock Centernet: Keypoint triplets for object detection.
\newblock In \emph{ICCV}, 2019.

\bibitem[Fang et~al.(2021)Fang, Yang, Wang, Li, Fang, Shan, Feng, and
  Liu]{Fang_2021_ICCV}
Fang, Y., Yang, S., Wang, X., Li, Y., Fang, C., Shan, Y., Feng, B., and Liu, W.
\newblock Instances as queries.
\newblock In \emph{ICCV}, 2021.

\bibitem[Gao et~al.(2021)Gao, Zheng, Wang, Dai, and Li]{gao2021fast}
Gao, P., Zheng, M., Wang, X., Dai, J., and Li, H.
\newblock Fast convergence of detr with spatially modulated co-attention.
\newblock In \emph{ICCV}, 2021.

\bibitem[Girshick(2015)]{girshick2015fast}
Girshick, R.
\newblock Fast r-cnn.
\newblock In \emph{ICCV}, 2015.

\bibitem[Han et~al.(2021)Han, Huang, Song, Yang, Wang, and
  Wang]{han2021dynamic}
Han, Y., Huang, G., Song, S., Yang, L., Wang, H., and Wang, Y.
\newblock Dynamic neural networks: A survey.
\newblock \emph{IEEE Transactions on Pattern Analysis and Machine
  Intelligence}, 2021.

\bibitem[He et~al.(2016)He, Zhang, Ren, and Sun]{he2016deep}
He, K., Zhang, X., Ren, S., and Sun, J.
\newblock Deep residual learning for image recognition.
\newblock In \emph{CVPR}, 2016.

\bibitem[He et~al.(2017)He, Gkioxari, Doll{\'a}r, and Girshick]{he2017mask}
He, K., Gkioxari, G., Doll{\'a}r, P., and Girshick, R.
\newblock Mask r-cnn.
\newblock In \emph{ICCV}, 2017.

\bibitem[Hosang et~al.(2017)Hosang, Benenson, and Schiele]{hosang2017learning}
Hosang, J., Benenson, R., and Schiele, B.
\newblock Learning non-maximum suppression.
\newblock In \emph{CVPR}, 2017.

\bibitem[Hu et~al.(2021)Hu, Cao, Lu, Zhang, Wang, Li, Huang, Shao, and
  Ji]{hu2021istr}
Hu, J., Cao, L., Lu, Y., Zhang, S., Wang, Y., Li, K., Huang, F., Shao, L., and
  Ji, R.
\newblock Istr: End-to-end instance segmentation with transformers.
\newblock \emph{arXiv preprint arXiv:2105.00637}, 2021.

\bibitem[Hwang et~al.(2021)Hwang, Heo, Oh, and Kim]{hwang2021video}
Hwang, S., Heo, M., Oh, S.~W., and Kim, S.~J.
\newblock Video instance segmentation using inter-frame communication
  transformers.
\newblock \emph{NIPS}, 2021.

\bibitem[Jain et~al.(2022)Jain, Li, Chiu, Hassani, Orlov, and
  Shi]{jain2022oneformer}
Jain, J., Li, J., Chiu, M., Hassani, A., Orlov, N., and Shi, H.
\newblock Oneformer: One transformer to rule universal image segmentation.
\newblock \emph{arXiv preprint arXiv:2211.06220}, 2022.

\bibitem[Jia et~al.(2022)Jia, Yuan, He, Wu, Yu, Lin, Sun, Zhang, and
  Hu]{jia2022detrs}
Jia, D., Yuan, Y., He, H., Wu, X., Yu, H., Lin, W., Sun, L., Zhang, C., and Hu,
  H.
\newblock Detrs with hybrid matching.
\newblock \emph{arXiv preprint arXiv:2207.13080}, 2022.

\bibitem[Kirillov et~al.(2019)Kirillov, He, Girshick, Rother, and
  Doll{\'a}r]{kirillov2019panoptic}
Kirillov, A., He, K., Girshick, R., Rother, C., and Doll{\'a}r, P.
\newblock Panoptic segmentation.
\newblock In \emph{CVPR}, 2019.

\bibitem[Li et~al.(2021)Li, Wang, Wang, Liang, Li, and Chang]{li2021dynamic}
Li, C., Wang, G., Wang, B., Liang, X., Li, Z., and Chang, X.
\newblock Dynamic slimmable network.
\newblock In \emph{CVPR}, 2021.

\bibitem[Li et~al.(2022)Li, Zhang, Liu, Guo, Ni, and Zhang]{li2022dn}
Li, F., Zhang, H., Liu, S., Guo, J., Ni, L.~M., and Zhang, L.
\newblock Dn-detr: Accelerate detr training by introducing query denoising.
\newblock In \emph{CVPR}, 2022.

\bibitem[Liang et~al.(2023)Liang, Zhou, Liu, and Wang]{liang2023clustseg}
Liang, J., Zhou, T., Liu, D., and Wang, W.
\newblock Clustseg: Clustering for universal segmentation.
\newblock \emph{arXiv preprint arXiv:2305.02187}, 2023.

\bibitem[Lin et~al.(2014)Lin, Maire, Belongie, Hays, Perona, Ramanan,
  Doll{\'a}r, and Zitnick]{lin2014microsoft}
Lin, T.-Y., Maire, M., Belongie, S., Hays, J., Perona, P., Ramanan, D.,
  Doll{\'a}r, P., and Zitnick, C.~L.
\newblock Microsoft coco: Common objects in context.
\newblock In \emph{ECCV}, 2014.

\bibitem[Lin et~al.(2017{\natexlab{a}})Lin, Doll{\'a}r, Girshick, He,
  Hariharan, and Belongie]{lin2017feature}
Lin, T.-Y., Doll{\'a}r, P., Girshick, R., He, K., Hariharan, B., and Belongie,
  S.
\newblock Feature pyramid networks for object detection.
\newblock In \emph{CVPR}, 2017{\natexlab{a}}.

\bibitem[Lin et~al.(2017{\natexlab{b}})Lin, Goyal, Girshick, He, and
  Doll{\'a}r]{lin2017focal}
Lin, T.-Y., Goyal, P., Girshick, R., He, K., and Doll{\'a}r, P.
\newblock Focal loss for dense object detection.
\newblock In \emph{ICCV}, 2017{\natexlab{b}}.

\bibitem[Liu et~al.(2021{\natexlab{a}})Liu, Cui, Tan, and Chen]{liu2021sg}
Liu, D., Cui, Y., Tan, W., and Chen, Y.
\newblock Sg-net: Spatial granularity network for one-stage video instance
  segmentation.
\newblock In \emph{CVPR}, 2021{\natexlab{a}}.

\bibitem[Liu et~al.(2022)Liu, Li, Zhang, Yang, Qi, Su, Zhu, and
  Zhang]{liu2022dabdetr}
Liu, S., Li, F., Zhang, H., Yang, X., Qi, X., Su, H., Zhu, J., and Zhang, L.
\newblock {DAB}-{DETR}: Dynamic anchor boxes are better queries for {DETR}.
\newblock In \emph{ICLR}, 2022.

\bibitem[Liu et~al.(2019)Liu, Ren, and Ye]{liu2019spatio}
Liu, X., Ren, H., and Ye, T.
\newblock Spatio-temporal attention network for video instance segmentation.
\newblock In \emph{ICCVW}, 2019.

\bibitem[Liu et~al.(2021{\natexlab{b}})Liu, Lin, Cao, Hu, Wei, Zhang, Lin, and
  Guo]{liu2021swin}
Liu, Z., Lin, Y., Cao, Y., Hu, H., Wei, Y., Zhang, Z., Lin, S., and Guo, B.
\newblock Swin transformer: Hierarchical vision transformer using shifted
  windows.
\newblock In \emph{ICCV}, 2021{\natexlab{b}}.

\bibitem[Meng et~al.(2021)Meng, Chen, Fan, Zeng, Li, Yuan, Sun, and
  Wang]{meng2021-CondDETR}
Meng, D., Chen, X., Fan, Z., Zeng, G., Li, H., Yuan, Y., Sun, L., and Wang, J.
\newblock Conditional detr for fast training convergence.
\newblock In \emph{ICCV}, 2021.

\bibitem[Neubeck \& Van~Gool(2006)Neubeck and Van~Gool]{neubeck2006efficient}
Neubeck, A. and Van~Gool, L.
\newblock Efficient non-maximum suppression.
\newblock In \emph{ICPR}, 2006.

\bibitem[Ren et~al.(2015)Ren, He, Girshick, and Sun]{ren2015faster}
Ren, S., He, K., Girshick, R., and Sun, J.
\newblock Faster r-cnn: Towards real-time object detection with region proposal
  networks.
\newblock In \emph{NIPS}, 2015.

\bibitem[Roh et~al.(2021)Roh, Shin, Shin, and Kim]{roh2021sparse}
Roh, B., Shin, J., Shin, W., and Kim, S.
\newblock Sparse detr: Efficient end-to-end object detection with learnable
  sparsity.
\newblock \emph{arXiv preprint arXiv:2111.14330}, 2021.

\bibitem[Rothe et~al.(2015)Rothe, Guillaumin, and Van~Gool]{rothe2015non}
Rothe, R., Guillaumin, M., and Van~Gool, L.
\newblock Non-maximum suppression for object detection by passing messages
  between windows.
\newblock In \emph{ACCV}, 2015.

\bibitem[Thawakar et~al.(2022)Thawakar, Narayan, Cao, Cholakkal, Anwer, Khan,
  Khan, Felsberg, and Khan]{thawakar2022video}
Thawakar, O., Narayan, S., Cao, J., Cholakkal, H., Anwer, R.~M., Khan, M.~H.,
  Khan, S., Felsberg, M., and Khan, F.~S.
\newblock Video instance segmentation via multi-scale spatio-temporal split
  attention transformer.
\newblock In \emph{ECCV}, 2022.

\bibitem[Tian et~al.(2020)Tian, Shen, and Chen]{tian2020conditional}
Tian, Z., Shen, C., and Chen, H.
\newblock Conditional convolutions for instance segmentation.
\newblock In \emph{ECCV}, 2020.

\bibitem[Van~der Maaten \& Hinton(2008)Van~der Maaten and
  Hinton]{van2008visualizing}
Van~der Maaten, L. and Hinton, G.
\newblock Visualizing data using t-sne.
\newblock \emph{Journal of machine learning research}, 9\penalty0 (11), 2008.

\bibitem[Vaswani et~al.(2017)Vaswani, Shazeer, Parmar, Uszkoreit, Jones, Gomez,
  Kaiser, and Polosukhin]{vaswani2017attention}
Vaswani, A., Shazeer, N., Parmar, N., Uszkoreit, J., Jones, L., Gomez, A.~N.,
  Kaiser, {\L}., and Polosukhin, I.
\newblock Attention is all you need.
\newblock \emph{NIPS}, 2017.

\bibitem[Wang et~al.(2021{\natexlab{a}})Wang, Zhu, Adam, Yuille, and
  Chen]{wang2021max}
Wang, H., Zhu, Y., Adam, H., Yuille, A., and Chen, L.-C.
\newblock Max-deeplab: End-to-end panoptic segmentation with mask transformers.
\newblock In \emph{CVPR}, 2021{\natexlab{a}}.

\bibitem[Wang et~al.(2021{\natexlab{b}})Wang, Liew, Li, Chen, and
  Feng]{wang2021sodar}
Wang, T., Liew, J.~H., Li, Y., Chen, Y., and Feng, J.
\newblock Sodar: Exploring locally aggregated learning of mask representations
  for instance segmentation.
\newblock \emph{IEEE Transactions on Image Processing}, 31:\penalty0 839--851,
  2021{\natexlab{b}}.

\bibitem[Wang et~al.(2020{\natexlab{a}})Wang, Kong, Shen, Jiang, and
  Li]{wang2020solo}
Wang, X., Kong, T., Shen, C., Jiang, Y., and Li, L.
\newblock {SOLO}: Segmenting objects by locations.
\newblock In \emph{ECCV}, 2020{\natexlab{a}}.

\bibitem[Wang et~al.(2020{\natexlab{b}})Wang, Zhang, Kong, Li, and
  Shen]{wang2020solov2}
Wang, X., Zhang, R., Kong, T., Li, L., and Shen, C.
\newblock Solov2: Dynamic and fast instance segmentation.
\newblock \emph{NIPS}, 2020{\natexlab{b}}.

\bibitem[Wang et~al.(2021{\natexlab{c}})Wang, Huang, Song, Huang, and
  Huang]{wang2021not}
Wang, Y., Huang, R., Song, S., Huang, Z., and Huang, G.
\newblock Not all images are worth 16x16 words: Dynamic transformers for
  efficient image recognition.
\newblock \emph{NIPS}, 2021{\natexlab{c}}.

\bibitem[Wang et~al.(2021{\natexlab{d}})Wang, Xu, Wang, Shen, Cheng, Shen, and
  Xia]{wang2021end}
Wang, Y., Xu, Z., Wang, X., Shen, C., Cheng, B., Shen, H., and Xia, H.
\newblock End-to-end video instance segmentation with transformers.
\newblock In \emph{CVPR}, 2021{\natexlab{d}}.

\bibitem[Wang et~al.(2022)Wang, Zhang, Yang, and Sun]{wang2022anchor}
Wang, Y., Zhang, X., Yang, T., and Sun, J.
\newblock Anchor detr: Query design for transformer-based detector.
\newblock In \emph{AAAI}, 2022.

\bibitem[Wu et~al.(2022)Wu, Jiang, Bai, Zhang, and Bai]{wu2022seqformer}
Wu, J., Jiang, Y., Bai, S., Zhang, W., and Bai, X.
\newblock Seqformer: Sequential transformer for video instance segmentation.
\newblock In \emph{ECCV}, 2022.

\bibitem[Xiong et~al.(2019)Xiong, Liao, Zhao, Hu, Bai, Yumer, and
  Urtasun]{xiong2019upsnet}
Xiong, Y., Liao, R., Zhao, H., Hu, R., Bai, M., Yumer, E., and Urtasun, R.
\newblock Upsnet: A unified panoptic segmentation network.
\newblock In \emph{CVPR}, 2019.

\bibitem[Yang et~al.(2019{\natexlab{a}})Yang, Bender, Le, and
  Ngiam]{yang2019condconv}
Yang, B., Bender, G., Le, Q.~V., and Ngiam, J.
\newblock Condconv: Conditionally parameterized convolutions for efficient
  inference.
\newblock \emph{NIPS}, 2019{\natexlab{a}}.

\bibitem[Yang et~al.(2019{\natexlab{b}})Yang, Fan, and Xu]{yang2019video}
Yang, L., Fan, Y., and Xu, N.
\newblock Video instance segmentation.
\newblock In \emph{ICCV}, 2019{\natexlab{b}}.

\bibitem[Yang et~al.(2022)Yang, Wang, Li, Fang, Fang, Liu, Zhao, and
  Shan]{yang2022temporally}
Yang, S., Wang, X., Li, Y., Fang, Y., Fang, J., Liu, W., Zhao, X., and Shan, Y.
\newblock Temporally efficient vision transformer for video instance
  segmentation.
\newblock In \emph{CVPR}, 2022.

\bibitem[Yu \& Huang(2019)Yu and Huang]{yu2019autoslim}
Yu, J. and Huang, T.
\newblock Autoslim: Towards one-shot architecture search for channel numbers.
\newblock \emph{arXiv preprint arXiv:1903.11728}, 2019.

\bibitem[Yu et~al.(2018)Yu, Yang, Xu, Yang, and Huang]{yu2018slimmable}
Yu, J., Yang, L., Xu, N., Yang, J., and Huang, T.
\newblock Slimmable neural networks.
\newblock \emph{arXiv preprint arXiv:1812.08928}, 2018.

\bibitem[Zhang et~al.(2020)Zhang, Chang, Ma, Wang, and Chen]{zhang2020dynamic}
Zhang, H., Chang, H., Ma, B., Wang, N., and Chen, X.
\newblock Dynamic r-cnn: Towards high quality object detection via dynamic
  training.
\newblock In \emph{ECCV}, 2020.

\bibitem[Zhang et~al.(2022)Zhang, Li, Liu, Zhang, Su, Zhu, Ni, and
  Shum]{zhang2022dino}
Zhang, H., Li, F., Liu, S., Zhang, L., Su, H., Zhu, J., Ni, L.~M., and Shum,
  H.-Y.
\newblock Dino: Detr with improved denoising anchor boxes for end-to-end object
  detection.
\newblock \emph{arXiv preprint arXiv:2203.03605}, 2022.

\bibitem[Zhang et~al.(2021)Zhang, Pang, Chen, and Loy]{zhang2021k}
Zhang, W., Pang, J., Chen, K., and Loy, C.~C.
\newblock K-net: Towards unified image segmentation.
\newblock \emph{NIPS}, 2021.

\bibitem[Zhu et~al.(2021)Zhu, Su, Lu, Li, Wang, and Dai]{zhu2021deformable}
Zhu, X., Su, W., Lu, L., Li, B., Wang, X., and Dai, J.
\newblock Deformable detr: Deformable transformers for end-to-end object
  detection.
\newblock In \emph{ICLR}, 2021.

\end{thebibliography}
\bibliographystyle{icml2023}

% %%%%%%%%%%%%%%%%%%%%%%%%%%%%%%%%%%%%%%%%%%%%%%%%%%%%%%%%%%%%%%%%%%%%%%%%%%%%%%%
% %%%%%%%%%%%%%%%%%%%%%%%%%%%%%%%%%%%%%%%%%%%%%%%%%%%%%%%%%%%%%%%%%%%%%%%%%%%%%%%
% % APPENDIX
% %%%%%%%%%%%%%%%%%%%%%%%%%%%%%%%%%%%%%%%%%%%%%%%%%%%%%%%%%%%%%%%%%%%%%%%%%%%%%%%
% %%%%%%%%%%%%%%%%%%%%%%%%%%%%%%%%%%%%%%%%%%%%%%%%%%%%%%%%%%%%%%%%%%%%%%%%%%%%%%%
% \newpage
% \appendix
% \onecolumn
% \section{You \emph{can} have an appendix here.}

% You can have as much text here as you want. The main body must be at most $8$ pages long.
% For the final version, one more page can be added.
% If you want, you can use an appendix like this one, even using the one-column format.
% %%%%%%%%%%%%%%%%%%%%%%%%%%%%%%%%%%%%%%%%%%%%%%%%%%%%%%%%%%%%%%%%%%%%%%%%%%%%%%%
% %%%%%%%%%%%%%%%%%%%%%%%%%%%%%%%%%%%%%%%%%%%%%%%%%%%%%%%%%%%%%%%%%%%%%%%%%%%%%%%

\end{document}